\documentclass[letterpaper]{article} 
\usepackage{aaai2026}  
\usepackage{times}  
\usepackage{helvet}  
\usepackage{courier}  
\usepackage[hyphens]{url}  
\usepackage{graphicx} 
\usepackage{natbib}  
\usepackage{caption} 

\usepackage{booktabs} 
\usepackage{multirow}
\usepackage{amsmath} 
\usepackage{amsfonts}
\usepackage{soul}

\usepackage{siunitx}
\usepackage[table,xcdraw]{xcolor}
\usepackage{colortbl}             

\usepackage{algorithm}
\usepackage{algorithmic}

\usepackage{newfloat}
\usepackage{listings}
\DeclareCaptionStyle{ruled}{labelfont=normalfont,labelsep=colon,strut=off} 
\floatstyle{ruled}
\newfloat{listing}{tb}{lst}{}
\floatname{listing}{Listing}
\title{Pb4U-GNet: Resolution-Adaptive Garment Simulation via Propagation-before-Update Graph Network}
\author{
    Aoran Liu\textsuperscript{\rm 1},
    Kun Hu\textsuperscript{\rm 2,}\thanks{Corresponding author.},
    Clinton Ansun Mo\textsuperscript{\rm 3}, 
    Qiuxia Wu\textsuperscript{\rm 4}, 
    Wenxiong Kang\textsuperscript{\rm 5}, 
    Zhiyong Wang\textsuperscript{\rm 1}
}
\affiliations{
    \textsuperscript{\rm 1}School of Computer Science, The University of Sydney, Camperdown, NSW, Australia\\
    \textsuperscript{\rm 2}School of Science, Edith Cowan University, Joondalup, WA, Australia\\
    \textsuperscript{\rm 3}Matsuo-Iwasawa Lab, The University of Tokyo, Tokyo, Japan\\
    \textsuperscript{\rm 4}School of Software Engineering, South China University of Technology, Guangzhou, China\\
    \textsuperscript{\rm 5}School of Automation Science \& Engineering, South China University of Technology, Guangzhou, China\\

    aliu4429@uni.sydney.edu.au, k.hu@ecu.edu.au, clinton.mo@weblab.t.u-tokyo.ac.jp,\\ \{qxwu, auwxkang\}@scut.edu.cn, zhiyong.wang@sydney.edu.au
}

\usepackage{bibentry}

\begin{document}

\maketitle

\begin{abstract}
Garment simulation is fundamental to various applications in computer vision and graphics, from virtual try-on to digital human modelling. However, conventional physics-based methods remain computationally expensive, hindering their application in time-sensitive scenarios. While graph neural networks (GNNs) offer promising acceleration, existing approaches exhibit poor cross-resolution generalisation, demonstrating significant performance degradation on higher-resolution meshes beyond the training distribution. This stems from two key factors: (1) existing GNNs employ fixed message-passing depth that fails to adapt information aggregation to mesh density variation, and (2) vertex-wise displacement magnitudes are inherently resolution-dependent in garment simulation.
To address these issues, we introduce Propagation-before-Update Graph Network (Pb4U-GNet), a resolution-adaptive framework that decouples message propagation from feature updates. Pb4U-GNet incorporates two key mechanisms: (1) dynamic propagation depth control, adjusting message-passing iterations based on mesh resolution, and (2) geometry-aware update scaling, which scales predictions according to local mesh characteristics. Extensive experiments show that even trained solely on low-resolution meshes, Pb4U-GNet exhibits strong generalisability across diverse mesh resolutions, addressing a fundamental challenge in neural garment simulation.
\end{abstract}

\begin{links}
    \link{Code}{https://github.com/adam-lau709/PB4U-GNet}
\end{links}

\section{Introduction}

Realistic cloth and garment simulation play a crucial role in many computer vision and graphics applications, including virtual try-on, virtual-reality experiences and digital human modelling. Conventional methods model cloth with physics-based formulations, such as mass-spring systems~\cite{provot1995deformation}, to approximate internal elastic forces and reproduce its dynamic behaviour. 
These approaches employ numerical integration to advance the simulation forward in time, while an iterative solver enforces physical constraints at each timestep to maintain equilibrium. However, for high-resolution meshes, these repeated constraint-solving iterations become computationally prohibitive, making physics-based techniques expensive for real-time applications.

\begin{figure}[h!]
  \includegraphics[width=\columnwidth]{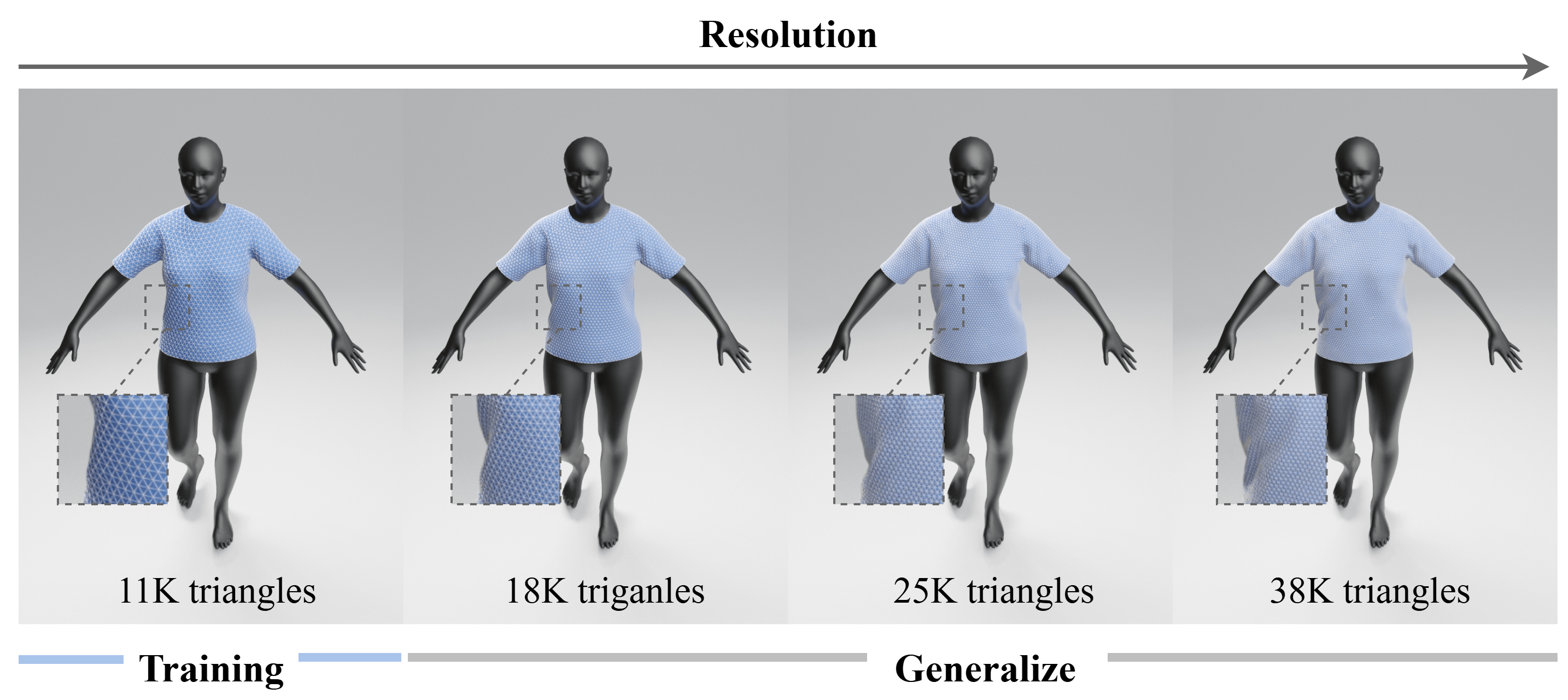}
  \caption{Sample results of Pb4U-GNet, a resolution-adaptive garment simulation framework based on graph neural networks. Trained on low resolution meshes with $\sim$11K triangles, Pb4U-GNet generalises effectively to significantly higher resolutions, producing stable and realistic simulation results without retraining.}
  \label{fig:teaser}
\end{figure}

To accelerate simulation, deep learning methods have been proposed as alternatives to physics-based solvers~\cite{gundogdu2019garnet, santesteban2019learning, pfaff2020learning}.
Among them, graph neural networks (GNNs)~\cite{scarselli2008graph} have emerged as powerful predictors of garment dynamics, combining strong generalisability with visually realistic results. 
GNNs perform message passing: every garment vertex exchanges state information with its neighbours, updates its latent feature, and then predicts its next position via a learnable function.

Despite their promise, current GNN-based garment simulators exhibit poor generalisation across different mesh resolutions. Models trained on a specific mesh resolution often fail on meshes with different densities, particularly those with higher resolution than encountered during training. This constrains practical deployment, where mesh resolution must adapt dynamically to varying computational budgets and application-specific requirements. Moreover, training directly on high-resolution meshes is often computationally prohibitive, creating a practical dilemma for achieving robust cross-resolution performance.



We identify two factors that drive this resolution-transfer failure. First, a fixed message-passing depth confines each vertex to a preset hop distance, so on a fine mesh the network sees too little context, while increasing the depth for a coarse mesh over-smooths the result. Second, deformation magnitude scales with mesh density: global motion is distributed across more vertices in finer meshes, leading to diminished per-vertex displacement. This mismatch introduces physically inconsistent predictions on unseen resolutions.

To address these challenges, we propose Propagation-before-Update Graph Network (Pb4U-GNet), a resolution-adaptive framework that decouples message propagation from feature updates: it first performs iterative message passing to flexibly control receptive fields, then updates all vertex features collectively.

Building on this design, we introduce two complementary strategies: (1) \textit{resolution-aware propagation control} that dynamically adjusts message-passing depth to maintain consistent receptive fields across resolutions, and (2) \textit{resolution-aware update scaling} that rescales predicted accelerations based on local geometric scale for physically consistent deformation.
Extensive evaluations demonstrate that Pb4U-GNet significantly outperforms existing methods in cross-resolution generalisation, even when trained exclusively on low-resolution inputs. Our key contributions are:

\begin{itemize}

\item We propose Pb4U-GNet, a novel Propagation-before-Update Graph Network for garment simulation, capable of generalising to unseen garment resolutions even when trained solely on low-resolution meshes.

\item We introduce a resolution-aware propagation control strategy that dynamically adjusts the message passing depth based on mesh resolution, maintaining consistent spatial coverage across varying discretizations.

\item We devise a resolution-aware scaling update strategy that normalises the predicted vertex acceleration according to their geometric scale, ensuring physically consistent deformation across meshes of different densities.
\end{itemize}

\section{Related Work}


\noindent \textbf{Physics-Based Garment Simulation}. Garment simulation has been a long-standing challenge in computer graphics. Conventional physics-based methods use models such as mass-spring systems~\cite{provot1995deformation} to compute garment dynamics, with subsequent improvements in numerical stability~\cite{baraff1998large} and yarn-level modelling~\cite{kaldor2008simulating, kaldor2010efficient}. To address computational costs, constraint-based approaches like Position-Based Dynamics~\cite{muller2007position} and Projective Dynamics~\cite{bouaziz2014projective, liu2013fast} have been developed for faster simulation. However, achieving efficient garment simulation while preserving realistic detail remains challenging.

\begin{figure*}
  \includegraphics[width=\textwidth]{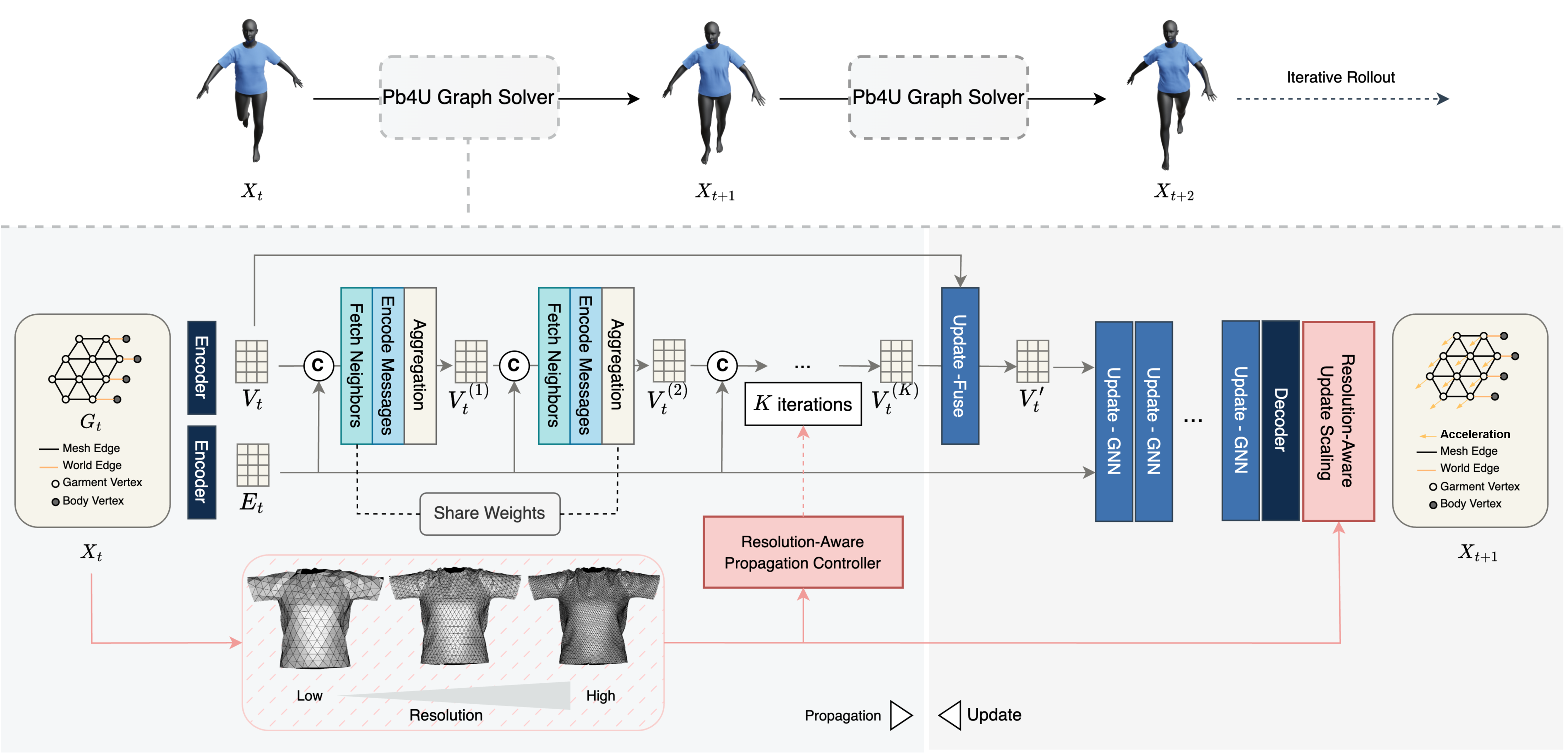}
  \caption{Illustration of the proposed proposed Pb4U-GNet, which decouples message propagation with a \textit{propagation-before-update} scheme.
With a \textit{resolution-aware propagation control} and a \textit{resolution-aware update scaling} design, it enables resolution-adaptive garment simulation. }
  \label{fig:one}
\end{figure*}

\noindent\textbf{Pose-Conditioned Garments Modelling}. Deep learning has emerged as an efficient alternative for garment simulation, with many methods predicting garment deformation conditioned on body pose using simulation data~\cite{wang2010example, santesteban2019learning, santesteban2021self, patel2020tailornet, gundogdu2019garnet, bertiche2021iccv} or 3D scans~\cite{saito2021scanimate, xiang2023drivable, lahner2018deepwrinkles, ponsmoll2017clothcap} as supervision. Recent approaches~\cite{bertiche2021pbns, santesteban2022snug, bertiche2022ncs} incorporate physics-based supervision by minimising the internal energy of the predicted garment, yielding more realistic wrinkles than purely data-driven methods. However, pose-conditioned models often learn a fixed mapping that generalises poorly and typically require separate models for each garment, limiting scalability to different geometries and materials.

\noindent\textbf{Vertex-Level Dynamics Learning with Graph Networks}. Graph Neural Networks (GNNs) have recently shown strong potential for modelling garment dynamics~\cite{sanchez2020learning, pfaff2020learning}, predicting per-vertex accelerations by propagating local features across the mesh. Their locality enables geometry-agnostic and generalisable behaviour. However, accurately capturing elastic wave propagation often requires deep message passing, which is computationally expensive on high-resolution meshes.


To improve scalability, recent work has introduced hierarchical graph structures \cite{grigorev2023hood, fortunato2022multiscalemeshgraphnets, nabian2024xmeshgraphnet, grigorev2024contourcraft} to enable efficient long-range interactions. However, a key challenge remains: existing models struggle to generalise across mesh resolutions, limiting their applicability in real-world scenarios where resolution can vary significantly. This highlights the need for a resolution-adaptive framework to ensure practical and scalable GNN-based garment simulation.

\noindent\textbf{Super-Resolution for Garment Simulation}. 
Super-resolution techniques aim to enhance low-resolution garment simulations by recovering high-resolution details. \cite{zhang2021deep} uses a CNN to refine garment normal maps, enhancing surface detail. \cite{halimi2023physgraph} proposes a physics-guided GNN to refine coarse simulation result. \cite{zhang2024neural} introduces a GNN-based framework with a neural interpolation scheme to propagate features from coarse to upsampled vertices. \cite{yu2024super} combines image-based super-resolution with a GNN module to enforce temporal coherence. However, these methods depend on a fixed-resolution coarse simulation. In contrast, our approach directly simulates garments at arbitrary resolutions without requiring a predefined mesh.


\section{Methodology}

As shown in Figure~\ref{fig:one}, the proposed Pb4U-GNet 
decouples message propagation from feature update in a \textit{propagation-before-update} scheme: it first performs iterative message passing to flexibly control the receptive field, then updates features collectively.
To enable resolution-adaptive modelling, a \textit{resolution-aware propagation control} and a \textit{resolution-aware update scaling} mechanisms are devised. 

\subsection{Graph Representation}

Let \( \mathbf{X}_g \in \mathbb{R}^{n_g\times3}\) and  \( \mathbf{X}_b \in \mathbb{R}^{n_b\times3} \) represent the garment and body mesh vertices, with \( n_g \) and \( n_b \) vertices, respectively. 
Garment simulation can be formulated as an autoregressive prediction problem over a sequence of mesh states \( \mathbf{X} = (\mathbf{X}_0, \mathbf{X}_1, \ldots, \mathbf{X}_n) \), where each state \( \mathbf{X}_t = (\mathbf{X}_{g,t}, \mathbf{X}_{b,t}, \mathcal{E}_t) \) represents the positions of both garment and body vertices at time step $t$. 
For vertex connectivity $\mathcal{E}_t$, in addition to the mesh topology, we follow the approach of \cite{pfaff2020learning} and incorporate garment-body interactions by introducing \emph{world edges} between garment and body vertices based on spatial proximity. Specifically, a world edge is added between a garment vertex $x_i \in \mathbf{X}_{g,t}$ and a body vertex $x_j \in \mathbf{X}_{b,t}$ if the Euclidean distance between them is below a predefined threshold.



The vertex features include physical and geometric attributes such as velocity, mass, surface normals, material parameters, and a vertex-type indicator (garment or body). Notably, absolute spatial information, such as vertex positions, is excluded to ensure invariance to global translation.
The edge features encode relative geometric relationships between connected vertices. These include: (1) the current relative direction vector between the two vertices; (2) the relative direction vector in the rest (undeformed) state; and (3) the relative edge length, defined as the ratio between the current  and rest length. To promote resolution-adaptive modelling, absolute edge lengths are avoided. Instead, all edge features are formulated in relative terms, enabling the model to generalise across garment meshes with varying resolutions and edge densities.

\subsection{Problem Formulation \& Pb4U-GNet}

Given the current mesh state \( \mathbf{X}_t \), 
the objective is to predict the garment deformation at the next time step, denoted as \( \hat{\mathbf{X}}_{g,t+1} \), which approximates the future garment state \( \mathbf{X}_{g,t+1} \). This prediction can be modelled by a function $f_\theta$ with learnable parameters in $\theta$: 
\begin{equation}
\hat{\mathbf{X}}_{g,\,t+1} = f_\theta(\mathbf{X}_{g,\,t},\, \mathbf{X}_{b,\,t}).
\label{eq:garment_prediction}
\end{equation}

Specifically, for $\mathbf{X}_t$, a vertex encoder first maps the raw vertex features into latent embeddings $\mathbf{V}_t$, while an edge encoder transforms edge features into latent embeddings $\mathbf{E}_t$. These encoded representations serve as the initial inputs to the message propagation and graph update. 
With the Pb4U design, each vertex aggregates information from its spatial and topological neighbours through message-passing. This process is performed iteratively for $K$ steps, producing intermediate embeddings $\mathbf{V}_{t}^\prime$. The number of message-passing steps $K$ is dynamically determined by a resolution-aware propagation control mechanism: for high-resolution meshes, $K$ is increased to ensure sufficient receptive field coverage; for low-resolution meshes, it is reduced to improve computational efficiency and prevent over-smoothing from irrelevant distant information.

The intermediate embeddings $\mathbf{V}_t^\prime$ are subsequently refined by a stack of conventional GNN layers~\cite{pfaff2020learning}, producing final latent representations $\mathbf{V}_t''$, which are then processed through a vertex decoder to predict vertex-wise garment accelerations $\tilde{\mathbf{A}}_{g,t}$. This acceleration is then scaled by a vertex-wise resolution-aware factor $\mathbf{S}$:

\begin{equation}
\mathbf{A}_{g,t} = \mathbf{S} \odot \tilde{\mathbf{A}}_{g,t}
\label{eq:scale}
\end{equation}

Finally, the garment mesh state at the next time step, $\hat{\mathbf{X}}_{g,t+1}$, is obtained by applying forward Euler integration. Specifically, the vertex velocities $\mathbf{U}_{g}$ are first updated as $\mathbf{U}_{g,t+1} = \mathbf{U}_{g,t} + \mathbf{A}_{g,t} \Delta t$, where $\Delta t$ is the time step size. Then, the vertex positions are computed as $\hat{\mathbf{X}}_{g,t+1} = \mathbf{X}_{g,t} + \mathbf{U}_{g,t+1} \Delta t$.

\subsection{Propagation-before-Update}


To support resolution-adaptive receptive fields, we propose a decoupled message-passing scheme that separates message propagation from feature updates. Given initial vertex and edge embeddings, messages are recursively accumulated over $K$ steps to expand the receptive field. A single update is then applied using the accumulated messages. This design allows $K$ to be adjusted by resolution: higher values for for high-resolution meshes to capture long-range dependencies, and smaller for coarse meshes to reduce overhead.


Formally, during the propagation stage, each vertex maintains an aggregated feature vector $\mathbf{h}_{t,i}$, which is initially set to the vertex feature embedding $\mathbf{v}_{t,i}$. At each aggregation step \( k \), the \( i^{\text{th}} \) garment vertex aggregates information from its neighbouring vertices and edges to compute an intermediate aggregated feature vector \( \tilde{\mathbf{h}}^k_{t,i} \), defined as:
\begin{equation}
\tilde{\mathbf{h}}^k_{t,i} = \text{LayerNorm}\left(\sum_{j \in \mathcal{N}(i)} f_m\left(\mathbf{h}^{k-1}_{t,i}, \mathbf{h}^{k-1}_{t,j}, \mathbf{e}_{t,ij}\right)\right),
\end{equation}
where 
$\mathbf{h}^{k-1}_{t,i}$ and $\mathbf{h}^{k-1}_{t,j}$ are the previous aggregated feature embeddings of the $i^{\text{th}}$ vertex and its neighbour; $\mathbf{e}_{t,ij}$ represents the edge features between them, and $\mathcal{N}(i)$ represents the set of vertex indices neighbouring the $i^{\text{th}}$ vertex. The learnable message propagation function $f_m(\cdot)$, implemented as a multi-layer perceptron (MLP), encodes input features from each neighbour into a message vector. These vectors are summed and then normalised using LayerNorm to produce the intermediate aggregated representation \( \tilde{\mathbf{h}}^k_{t,i} \). This is then combined with the previous aggregated feature vector via a decay-based accumulation:
\begin{equation}
\mathbf{h}^k_{t,i} = \gamma \cdot \mathbf{h}^{k-1}_{t,i} + \tilde{\mathbf{h}}^k_{t,i},
\end{equation}
where \(\gamma\) is a decay factor that controls the impact of earlier aggregated messages.

After \( K \) propagation steps, the accumulated message features for each vertex are combined with the original vertex embeddings using a learnable update function \( f_u \), which fuses the original and propagated features into a unified latent representation \( \mathbf{v}'_{t,i} \), defined as: 
\begin{equation}
\mathbf{v}'_{t,i} = f_u(\mathbf{v}_{t,i}, \mathbf{h}^K_{t,i}),
\end{equation}
where \( \mathbf{v}_{t,i} \) denotes the original vertex embedding of the $i^{\text{th}}$ vertex, and \( \textbf{h}^K_{t,i} \) is the aggregated feature vector after \( K \) iterations of message propagation. The update function \( f_u(\cdot) \) is also implemented as an MLP, and the set of updated vertex features for all garment vertices is denoted as \(\mathbf{V}'_t = \{ \mathbf{v}'_{t,i} \mid i \in \mathcal{V} \}\) where \( \mathcal{V} \) represents the set of all vertex indices.

The updated graph with \(\mathbf{V}'_t\) is further processed by a graph neural network (GNN) using a fixed number of layers, which refines the vertex features based on mesh connectivity and produces the final vertex embedding $\mathbf{V}_t''$. These final embeddings are then processed through a vertex decoder to predict vertex-wise garment accelerations.

By decoupling aggregation from update, we allow flexible control over the receptive field size without entangling it with update frequency, enabling better adaptation to a wide range of mesh topologies and resolutions.


\subsection{Resolution-Aware Propagation Control}

We define $D$ as the effective physical propagation distance, calibrated to the base resolution (i.e., the lowest-resolution meshes employed during training). Given $K_{base}$ propagation steps required for stable simulation at base resolution with mean edge length $\overline{L}_{base}$, we set
\begin{equation}
D = K_{\text{base}} \times \overline{L}_{\text{base}}.
\end{equation}
This defines a physically consistent propagation distance that remains invariant across resolutions. For meshes with mean edge length $\overline{L}$, the propagation steps are computed as
\begin{equation}
K = \lfloor D \times \bar{L}^{-1}\rfloor.
\end{equation}
As resolution increases (\(\bar{L}\) decreases), \(K\) increases proportionally, preserving a consistent physical receptive field.

\subsection{Resolution-Aware Update Scaling}

Although the decoupled message propagation mechanism allows flexible control of the receptive field, it does not inherently ensure resolution-adaptive predictions. This is because physical quantities such as displacement or acceleration depend on mesh resolution: in high-resolution meshes, each vertex represents a smaller area and mass, resulting in smaller per-vertex accelerations under the same global deformation. As a result, models trained on coarse meshes may overestimate displacements when applied to finer meshes.

Therefore, we introduce a resolution-aware update scaling mechanism based on per-vertex edge length. Motivated by geometric similarity principles in continuum mechanics, where displacement fields scale linearly with element size, we compute a vertex-specific scaling factor as the average length of its connected edges:
\begin{equation}
\mathbf{s}_{i} = \frac{1}{|\mathcal{N}(i)|}\sum_{j \in \mathcal{N}(i)} l_{ij},
\end{equation}
where $s_i \in \mathbf{S}$, which is used in formula \ref{eq:scale};
\(l_{ij}\) represents the Euclidean length of the edge connecting vertex \(i\) and the neighbour vertex with index \(j\) at rest state; $\mathcal{N}(i)$ represents the vertex indices neighbouring vertex $i$.
This scaling restores physically consistent displacement magnitudes by compensating for resolution-dependent geometric variation, enabling the model to learn resolution-invariant deformation behaviour and generalise across meshes of varying densities.

\subsection{Physics-Based Supervision}

Following \cite{grigorev2023hood}, we train our model in a fully self-supervised manner using six physics-based loss terms:
(1) Stretch loss $\mathcal{L}_\text{stretch}$
measures stretching and compression energy using the St. Venant–Kirchhoff model, encouraging the garment to maintain realistic material properties and avoid excessive deformation;
(2) Bending loss $\mathcal{L}_\text{bending}$
penalises curvature between adjacent mesh faces, promoting appropriate stiffness and preventing unnatural folding;
(3) Collision loss $\mathcal{L}_\text{collision}$
quantifies garment–body interpenetration as the sum of penetration depths across intersecting vertices, enforcing physical separation;
(4) Gravity loss $\mathcal{L}_\text{gravity}$
encourages natural draping by penalising vertically raised vertices to simulate gravity;
(5) Friction loss $\mathcal{L}_\text{friction}$
penalises tangential motion at garment–body contact points to reduce unrealistic sliding;
and (6) 
Inertia loss $\mathcal{L}_\text{inertia}$
promotes temporal coherence by penalising abrupt velocity changes across timesteps, preserving physical momentum. The composite loss function is:
\begin{align}
\mathcal{L} &= \mathcal{L}_\text{stretch} + \mathcal{L}_\text{bending} + \mathcal{L}_\text{collision} \notag \\
            &\quad + \mathcal{L}_\text{gravity} + \mathcal{L}_\text{friction} + \mathcal{L}_\text{inertia}.
\end{align}
This physics-based formulation enables flexible training across diverse motions and resolutions without ground-truth supervision.

\section{Experiments}

\subsection{Experimental Setup}

\noindent\textbf{Dataset}. We use the VTO dataset~\cite{santesteban2019learning}, which includes a diverse set of human motion sequences; four are held out for testing, and the remainder are used for training. The training set comprises four garment types (T-shirt, tank top, long-sleeve shirt, and long dress), each with five mesh resolutions ranging from 11K to 38K triangles. Only the lowest resolution is used for training, while higher resolutions are reserved for evaluation to assess resolution generalisation. 

\noindent\textbf{Evaluation Metric}. 
Since our method is self-supervised, we assess the physical plausibility of the simulations using the same physics-based loss terms applied during training following the same setting in existing studies \cite{grigorev2023hood}.

\noindent\textbf{Training}. 
During training, we randomly sample individual frames from the motion sequences in the training set. For each frame, the garment mesh is first deformed with linear blend skinning (LBS) \cite{santesteban2019learning} to match the corresponding SMPL body pose, providing a plausible but coarse initial state that our model subsequently refines. The model is trained for 100{,}000 iterations, which takes $\sim$36 hours on an NVIDIA RTX 4070 Ti GPU.

\noindent\textbf{Model Implementation}. 
Each vertex and edge in the input graph is first encoded to a 128-dimensional latent space, which serves as the initial feature representation. During message propagation, both the message and update functions are implemented as MLPs with two hidden layers of 128 units each. After propagation, the features are further processed by a GNN comprising 15 MeshGraphNet blocks \cite{pfaff2020learning}, which refine the aggregated vertex embeddings.

\subsection{Generalisation to Unseen Resolutions}

\textbf{Quantitative Evaluation}.
To evaluate the generalisation ability of our model across varying mesh complexities, we assess simulation accuracy on four garment resolutions, with the average triangle count ranging from level 1 (lowest) to level 4 (highest). 
The corresponding mesh resolutions are approximately: Level 1 - 11K triangles, Level 2 - 18K, Level 3 - 25K and Level 4 - 38K. All models are trained exclusively on garments with the lowest resolution (11K), while the other resolutions remain unseen during training. Evaluation is conducted on held-out motion sequences to ensure an unbiased assessment of generalisation performance.


\setlength{\tabcolsep}{0.7mm}
\begin{table}[h!]
\centering
\small
\begin{tabular}{l|ccccc}
\toprule
\textbf{Metric} & \textbf{MGN} & \textbf{HOOD} & \textbf{ESLR} & \textbf{CCRAFT} & \textbf{Ours} \\
\midrule
\multicolumn{6}{c}{\textbf{Lv.1 (11K)}} \\
\cmidrule{1-6}
Stretch   & 5.30E-02  & 6.83E-02  & 2.97E-02  & 1.22E-01  & 3.06E-02  \\
Bending   & 3.39E-03  & 2.34E-03  & 1.81E-03  & 2.96E-03  & 2.92E-03  \\
Collision & 1.71E-02  & 2.16E-02  & 1.99E-04  & 1.14E-06  & 1.62E-03  \\
Inertia   & 2.02E-03  & 1.80E-03  & 1.72E-03  & 1.91E-03  & 1.78E-03  \\
Gravity   & -7.20E-02 & -8.59E-02 & -6.01E-02 & -8.56E-02 & -5.49E-02 \\
Friction  & 1.25E-03  & 1.30E-03  & 1.09E-03  & 1.19E-03  & 1.34E-03  \\
\cmidrule{1-6}
Total     & 4.70E-03  & 9.45E-03  & \textbf{-2.56E-02} & 4.24E-02  & \ul{-1.66E-02} \\
\midrule

\multicolumn{6}{c}{\textbf{Lv.2 (18K)}} \\
\cmidrule{1-6}
Stretch   & 4.13E-01  & 3.56E-01  & 8.32E-02  & 1.89E-01  & 4.17E-02  \\
Bending   & 4.80E-03  & 4.67E-03  & 2.08E-03  & 4.12E-03  & 4.23E-03  \\
Collision & 1.12E-01  & 1.13E-04  & 5.60E-02  & 6.10E-04  & 1.21E-02  \\
Inertia   & 3.66E-03  & 2.83E-03  & 1.89E-03  & 1.88E-03  & 1.72E-03  \\
Gravity   & -1.02E-01 & -1.15E-01 & -8.38E-02 & -8.68E-02 & -5.30E-02 \\
Friction  & 1.36E-03  & 1.36E-03  & 1.17E-03  & 1.20E-03  & 1.43E-03  \\
\cmidrule{1-6}
Total     & 4.32E-01  & 2.49E-01  & \ul {6.06E-02}  & 1.10E-01  & \textbf{8.13E-03} \\
\midrule

\multicolumn{6}{c}{\textbf{Lv.3 (25K)}} \\
\cmidrule{1-6}
Stretch   & 1.43E+03  & 3.89E-01  & 2.27E-01  & 2.51E-01  & 5.70E-02  \\
Bending   & 1.23E-01  & 4.88E-03  & 2.52E-03  & 4.59E-03  & 5.83E-03  \\
Collision & 3.39E+00  & 2.40E-04  & 4.64E-02  & 4.37E-06  & 4.69E-02  \\
Inertia   & 1.08E-02  & 2.93E-03  & 2.10E-03  & 1.82E-03  & 1.70E-03  \\
Gravity   & -3.66E-01 & -1.20E-01 & -1.07E-01 & -8.68E-02 & -4.95E-02 \\
Friction  & 1.47E-03  & 1.35E-03  & 1.27E-03  & 1.22E-03  & 1.50E-03  \\
\cmidrule{1-6}
Total     & 1.44E+03  & 2.78E-01  & 1.73E-01  & \ul {1.72E-01}  & \textbf{6.34E-02} \\
\midrule


\multicolumn{6}{c}{\textbf{Lv.4 (38K)}} \\
\cmidrule{1-6}
Stretch   & 1.24E+06  & 2.52E+00  & 1.07E+05  & 3.60E-01  & 1.30E-01  \\
Bending   & 1.03E+01  & 1.56E-02  & 9.14E-01  & 6.13E-03  & 1.43E-02  \\
Collision & 8.22E+00  & 1.97E-01  & 2.46E+01  & 3.84E-04  & 1.03E-01  \\
Inertia   & 4.07E-02  & 5.08E-03  & 9.95E-03  & 1.87E-03  & 1.64E-03  \\
Gravity   & -2.55E+00 & -1.78E-01 & -3.25E-02 & -8.72E-02 & -2.81E-02 \\
Friction  & 1.70E-03  & 1.65E-03  & 1.66E-03  & 1.24E-03  & 1.68E-03  \\
\cmidrule{1-6}
Total     & 1.24E+06  & 2.57E+00  & 1.07E+05  & \ul {2.82E-01}  & \textbf{2.22E-01} \\
\bottomrule
\end{tabular}
\caption{Performance comparison with state-of-the-art methods across different mesh resolutions, evaluated using physics-based loss metrics.}
\label{tab:quantitative}
\end{table}

Table~\ref{tab:quantitative} summarises the evaluation results, reporting individual physics-based metrics along with the total loss. The physics loss metrics are treated as the residuals of the cloth energy minimisation during training. Lower values therefore indicate closer convergence to a physically valid state, providing a measure of physical plausibility. 
We compare our method against four state-of-the-art graph-based simulators: MGN~\cite{pfaff2020learning}, HOOD~\cite{grigorev2023hood}, ESLR~\cite{liu2025extendedshortlongrangemesh} and CCRAFT~\cite{grigorev2024contourcraft}. It can be observed that, at the lowest resolution (11K), our method performs comparably to existing approaches. However, as the resolution increases beyond the training level, it significantly outperforms the other methods. In particular, existing methods exhibit clear signs of divergence at high resolution, most notably in the stretch loss. These results highlight the robustness and scalability of our approach in handling diverse mesh resolutions, especially in challenging high-resolution scenarios.

\begin{figure}[h!]
  \includegraphics[width=\columnwidth]{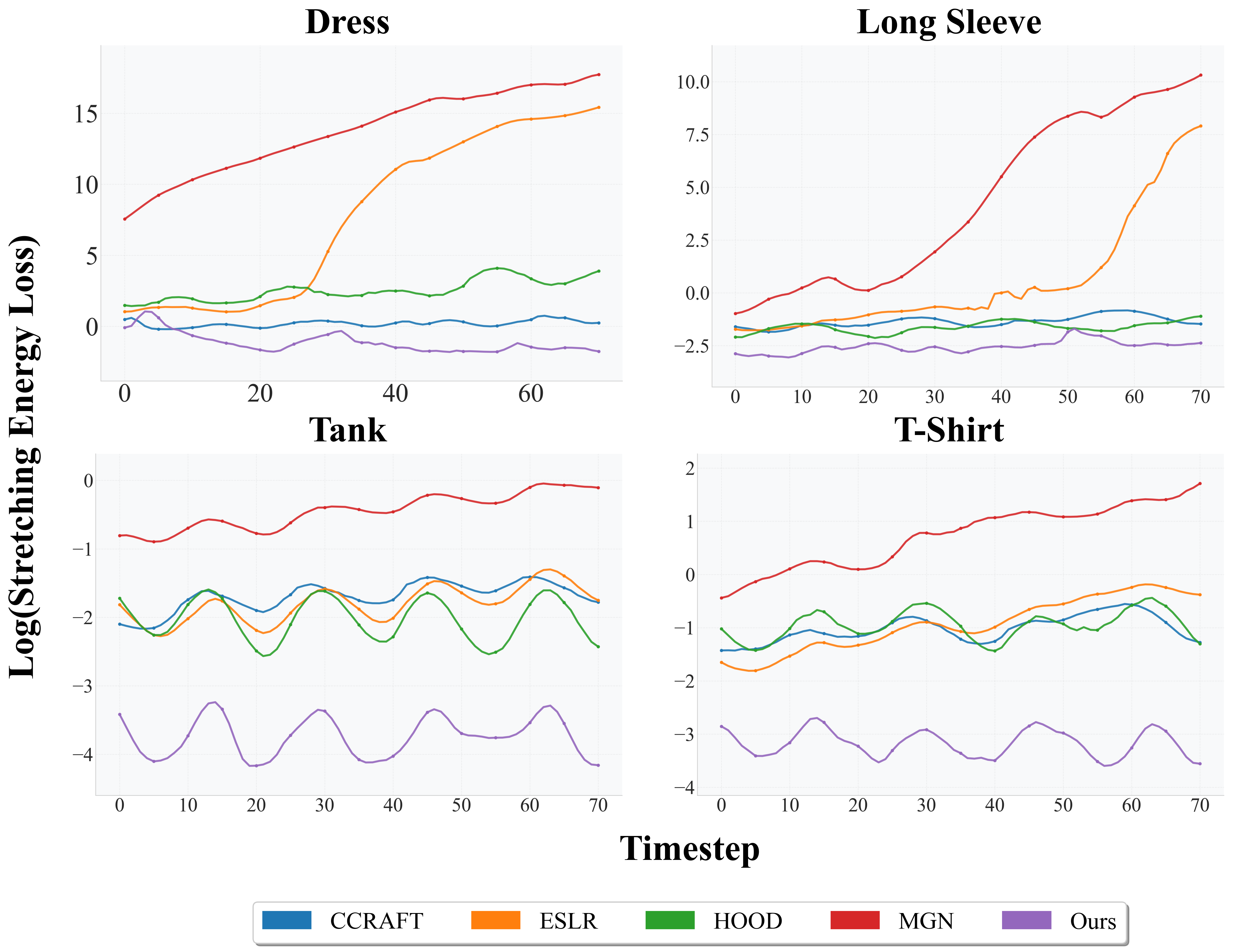}
  \caption{Stretch loss vs. time. The plots illustrate the temporal evolution of log stretching energy for each method on the test sequence 07\_02. Our method consistently maintains the lowest stretch energy across the simulations, demonstrating better physics validity.}
  \label{fig:line}
\end{figure}

Figure~\ref{fig:line} demonstrates the temporal evolution of the log stretching energy loss for each method. The simulations are conducted on the test sequence 07\_02, covering four different garment types under the highest resolution setting (38K). Low and stable stretch energy is essential for realistic garment simulation as it ensures the fabric behaves like real cloth, maintaining its original shape without unnatural stretching or shrinking over time. As shown, MGN and ESLR exhibit exploding stretch energy when simulating the dress and long-sleeve. Other methods remain more stable, while ours consistently achieves the lowest stretching energy across all garments, highlighting its robustness at high resolution.

\begin{figure}[h!]
  \includegraphics[width=\columnwidth]{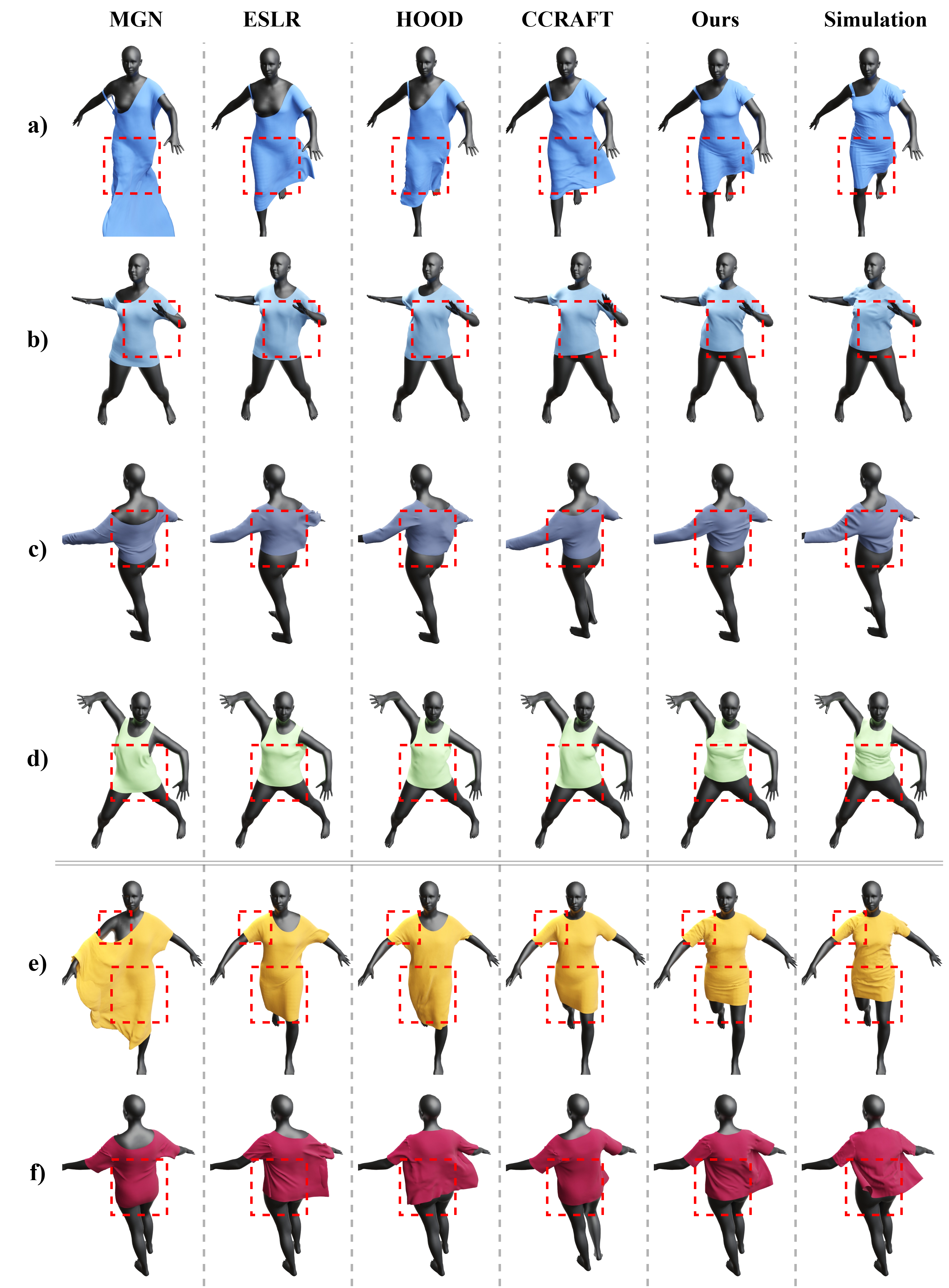}
  \caption{Rendered simulation results on high-resolution garment meshes. Baseline methods often struggle to preserve realistic fabric stretch, leading to noticeable over-stretched artefacts. Baseline methods also fail to preserve realistic wrinkle details.}
  \label{fig:qualitative}
\end{figure}

\noindent \textbf{Qualitative Evaluation}. 
Rows (a)–(d) of Figure~\ref{fig:qualitative} show visual comparisons of different methods on various motion frames using trained garment types at the highest resolution setting (38K). Physics-based simulation results are included as reference. Existing methods often struggle to preserve realistic garment behaviour: MGN shows severe distortions and topological artefacts (e.g., tearing, collapsing); HOOD and ESLR preserve the structure better but often exhibit slipping and misalignment, especially around the shoulders; CCRAFT performs best among baselines but still produces overstretched garments, lacking fine details like wrinkles.

In contrast, our method consistently generates realistic, physically plausible deformations that closely match the physics simulated results. Garments remain well-fitted, structurally intact, and exhibit realistic wrinkles, demonstrating strong robustness and generalisation under high-resolution, challenging scenarios.

\subsection{Generalisation to Unseen Garments}

\setlength{\tabcolsep}{0.7mm}
\begin{table}[h!]
\centering
\small
\begin{tabular}{lccccc}
\toprule
\textbf{Metric}   & \textbf{MGN} & \textbf{ESLR} & \textbf{HOOD} & \textbf{CCRAFT} & \textbf{Ours} \\
\midrule
Stretch   & 7.16E+04  & 1.49E+00  & 1.87E+00  & 3.11E-01  & 1.05E-01 \\
Bending   & 8.63E+00  & 1.23E-02  & 1.77E-02  & 1.03E-02  & 2.13E-02 \\
Collision & 3.40E+01  & 8.03E-05  & 2.06E-04  & 4.63E-04  & 3.05E-02 \\
Inertia   & 3.80E-02  & 2.15E-03  & 2.57E-03  & 1.88E-03  & 1.57E-03 \\
Gravity   & -2.08E+00 & -1.21E-01 & -1.52E-01 & -6.16E-02 & -5.33E-03 \\
Friction  & 2.52E-03  & 1.96E-03  & 1.75E-03  & 1.79E-03  & 2.33E-03 \\
\cmidrule{1-6}
Total     & 7.16E+04  & 1.39E+00  & 1.74E+00  & \ul{2.64E-01}  & \textbf{1.55E-01} \\
\bottomrule
\end{tabular}
\caption{Quantitative evaluation on unseen garments under the highest resolution setting.}
\label{tab:unseen-garments}
\end{table}

To assess the generalisability of our model on unseen garment categories, we evaluated two novel garment types: a form-fitting dress and a cardigan. All evaluations were conducted at the highest resolution. Table~\ref{tab:unseen-garments} shows the quantitative comparison between our method and state-of-the-art approaches. Our method achieves the lowest physics loss, confirming its robust generalisability to novel garments.

We present the rendered outputs for both garments in rows (e) and (f) of Figure~\ref{fig:qualitative}. Our method demonstrates better alignment with the physics simulation result, preserving realistic wrinkles and fabric dynamics. While competing methods display over-stretching and misalignment artefacts, as observed in the previous qualitative results.

\subsection{Propagation Depth Matters: Adapting to Mesh Complexity}


\begin{figure}[h!]
  \includegraphics[width=\columnwidth]{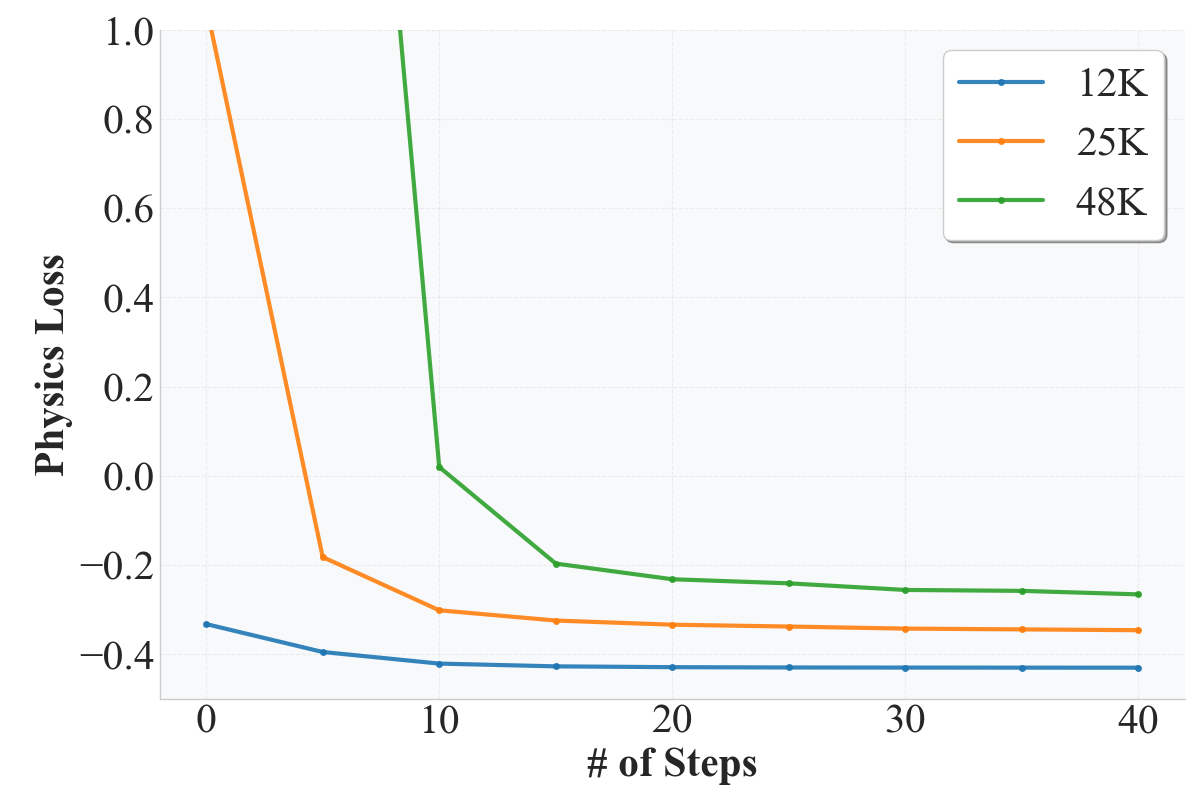}
  \caption{Physics loss vs. number of message propagation steps. Garments with different mesh resolutions require varying numbers of propagation steps to achieve stable and accurate simulation results.}
  \label{fig:acc-step}
\end{figure}

Figure~\ref{fig:acc-step} shows how total physics loss varies with the number of message propagation steps across three mesh resolutions (12K, 25K, 48K) using a dress template. Physics loss generally decreases with more steps, with higher-resolution meshes requiring more to converge, highlighting the link between resolution and receptive field size. Finer meshes need broader receptive fields to capture long-range interactions and maintain physical accuracy. Our method addresses this by adjusting the number of steps based on resolution, improving both efficiency and scalability.

In Table~\ref{tab:efficiency}, we present the runtime efficiency of our method compared to baseline models that use a fixed number of message propagation steps. For low-resolution meshes, our method adaptively reduces the number of propagation steps, resulting in improved computational efficiency while maintaining comparable simulation accuracy. For higher-resolution meshes, the model increases the number of steps as needed to preserve accuracy, achieving the lowest physics loss among all methods. These results highlight the strength of our adaptive propagation module, which dynamically allocates computation based on mesh complexity and incurs additional cost only when necessary.

\begin{table}[]
\centering
{%
\small          
\setlength{\tabcolsep}{3pt} 
\begin{tabular}{l@{\hskip 6pt}|l@{\hskip 6pt}|cc}
\toprule
\textbf{Resolution} & \textbf{Model} & \textbf{Physics Loss} & \textbf{Latency (ms)} \\
\midrule

\multirow{5}{*}{12K}
& MGN             & -4.22E-01          & \textbf{46.4} \\
& HOOD            & \textbf{-4.38E-01} & 50.8 \\
& ESLR            & -4.33E-01          & 54.6 \\
& CCRAFT          & -3.53E-01          & 97.2 \\
& \textbf{Pb4U-GNet (Ours)} & \ul {-4.11E-01}     & \ul {50.0} \\

\midrule
\multirow{5}{*}{25K}
& MGN             & -3.65E-01          & \ul {81.8} \\
& HOOD            & \ul {-3.81E-01}     & \textbf{81.5} \\
& ESLR            & -4.11E-01          & 98.2 \\
& CCRAFT          & -2.10E-01          & 171.4 \\
& \textbf{Pb4U-GNet (Ours)} & \textbf{-3.84E-01} & 105.3 \\

\midrule
\multirow{5}{*}{48K}
& MGN             & 5.75E+03           & \ul {144.5} \\
& HOOD            & 6.40E-01           & \textbf{141.1} \\
& ESLR            & \ul {1.38E-01}      & 184.8 \\
& CCRAFT          & 2.13E-01           & 761{,}499.3 \\
& \textbf{Pb4U-GNet (Ours)} & \textbf{-2.42E-01} & 196.4 \\

\bottomrule
\end{tabular}
}
\caption{Inference efficiency vs.\ simulation accuracy across mesh resolutions.}
\label{tab:efficiency}
\end{table}

\subsection{Ablation Study}

We conduct an ablation study to evaluate resolution-aware propagation control and update scaling modules. Table~\ref{tab:ablation} reports total physics loss across 11K, 25K, and 38K mesh resolutions. Removing propagation control maintains stable performance at 11K but causes sharp degradation at higher resolutions, showing that fixed receptive fields fail to capture long-range dependencies in finer meshes. Similarly, disabling update scaling significantly degrades performance at 25K and 38K, confirming the need to adapt predicted dynamics to mesh resolution. Both components are essential for robust, resolution-adaptive simulation.


\setlength{\tabcolsep}{1mm}
\begin{table}[h!]
\centering
\small
  \begin{tabular}{lrrr}
    \toprule
    \textbf{Model}                   & \textbf{Lv.1 (11K)}     & \textbf{Lv.3 (25K)}    & \textbf{Lv.4 (38K)} \\
    \midrule
    \textbf{Pb4U-GNet (Ours)}        & \textbf{-1.66E-02} & \textbf{6.34E-02} & \textbf{2.22E-01} \\
    w/o Propagation Control & -1.61E-03          & 1.08E+06          & 1.08E+09          \\
    w/o Update Scaling      & -5.78E-03          & 1.55E+13          & 7.34E+13          \\
    w/o Both                & 4.70E-03           & 1.44E+03          & 1.24E+06          \\
    \bottomrule
  \end{tabular}
    \caption{%
    Ablation study. }
\label{tab:ablation}
\end{table}

\section{Conclusion}
We present Pb4U-GNet, a framework for resolution-adaptive garment simulation. It incorporates a resolution-aware propagation control module that adjusts the message-passing depth based on mesh density, and an update scaling mechanism that modulates the model’s predictions according to mesh resolution. Experimental results demonstrate that Pb4U-GNet achieves superior accuracy and generalisation across a wide range of mesh resolutions.

\section{Acknowledgments}
This work was supported by the Australian Research Council (ARC) Linkage Project \#LP230100294 and ECU Science Early Career and New Staff Grant Scheme.


\bibliography{aaai2026}

\end{document}